\documentclass{article}

\usepackage[nonatbib,final]{neurips_2021}
\usepackage[utf8]{inputenc} 
\usepackage[T1]{fontenc}    
\usepackage{hyperref}       
\usepackage{url}            
\usepackage{booktabs}       
\usepackage{amsfonts}       
\usepackage{nicefrac}       
\usepackage{microtype}      

\usepackage{graphicx}
\usepackage{array}

\title{TagLab: A human-centric AI system for interactive semantic segmentation}

\author{%
  Gaia Pavoni \\
  Visual Computing Lab\\
  ISTI-CNR\\
  Pisa, Italy \\
  \texttt{gaia.pavoni@isti.cnr.it} \\
  \And
  Massimiliano Corsini\thanks{massimiliano.corsini@isti.cnr.it} \\
  Visual Computing Lab\\ 
  ISTI-CNR \\
  Pisa, Italy \\
  \texttt{massimiliano.corsini@isti.cnr.it} \\
  \And
  Federico Ponchio \\
  Visual Computing Lab \\
  ISTI-CNR \\
  Pisa, Italy \\
  \texttt{federico.ponchio@isti.cnr.it} \\
  \And
  Alessandro Muntoni  \\
  Visual Computing Lab \\
  ISTI-CNR \\
  Pisa, Italy \\
  \texttt{alessandro.muntoni@isti.cnr.it} \\
  \And
  Paolo Cignoni \\
  Visual Computing Lab \\
  ISTI-CNR \\
  Pisa, Italy \\
  \texttt{paolo.cignoni@isti.cnr.it} \\
}

\begin{document}

\maketitle

\begin{abstract}

Fully automatic semantic segmentation of highly specific semantic classes and complex shapes may not meet the accuracy standards demanded by scientists. In such cases, human-centered AI solutions, able to assist operators while preserving human control over complex tasks, are a good trade-off to speed up image labeling while maintaining high accuracy levels.
TagLab is an open-source AI-assisted software for annotating large orthoimages which takes advantage of different degrees of automation; it speeds up image annotation from scratch through assisted tools, creates custom fully automatic semantic segmentation models, and, finally, allows the quick edits of automatic predictions. Since the orthoimages analysis applies to several scientific disciplines, TagLab has been designed with a flexible labeling pipeline. We report our results in two different scenarios, marine ecology, and architectural heritage.

\end{abstract}

\section{Introduction}

Semantic segmentation is an interdisciplinary image analysis task, widely adopted for indoor scene understanding, autonomous driving, medical imaging, or satellite image data processing. In recent years, this task has been successfully automated by convolutional neural networks (CNNs). However, in some scientific applications where humans have to perform complex recognition tasks on complex data, the ability of experts to perform difficult choices remains irreplaceable.

In this context, we developed TagLab (\url{http://taglab.isti.cnr.it}), an open-source software tool for the AI-assisted semantic segmentation of orthoimages. TagLab's human-centered pipeline is \emph{flexible} as the workflow is not rigidly constrained, and \emph{effective} in reducing annotation time. Furthermore, the learning process, which allows for the creation of custom per-pixel classifiers, is designed for non-machine learning experts, making AI accessible to scientists with different backgrounds.

TagLab's development originally started under the pressure of a real challenge: monitoring corals reefs following the global warming threat. However, its design fits different application fields. 
This paper first describes the TagLab human-centered pipeline, comparing it to other workflows adopted by marine image data annotation systems. Then, we reports our experimental results in two applications: underwater monitoring and structural architecture.

\section{A human-centric AI pipeline for interactive regions labelling}

\begin{figure}
  \centering
  \includegraphics[width=1\textwidth]{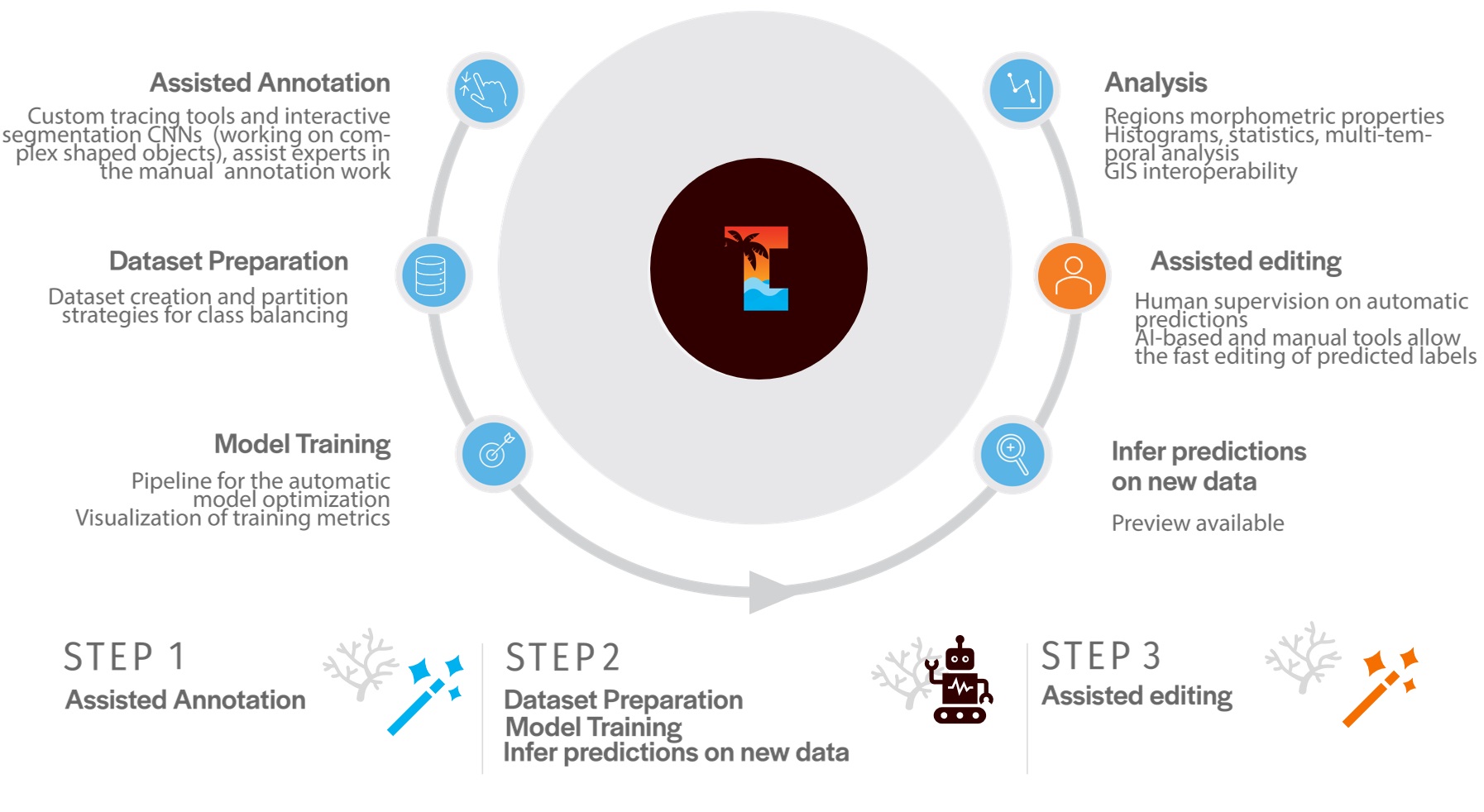}
  \caption{The proposed human-centric annotation pipeline. AI-based tools support the user in the annotation process (Step 1) and in the editing of automatic predictions (Step 3). The learning pipeline (Step 2) allows the creation of custom per-pixel classifiers.}
  \label{fig:pipeline}
\end{figure}

TagLab's human-centered AI pipeline, illustrated in Fig.\ref{fig:pipeline}, includes three steps:
\begin {description}
\item[Step 1: Assisted Annotation] {A combination of specialized and AI-based interactive tools speeds up the manual labeling.}
\item[Step 2: Learning pipeline] {After completing the annotation, the user is guided through the optimization of a fully automatic model. The custom pixel-classifier can be used to automate the annotation on other data belonging to the same domain.}
\item[Step 3: Assisted Editing] {The human expert re-enters the annotation cycle and corrects erroneous predictions using interactive tools, as in Step 1. Eventually, new labeled areas can be analyzed or included in a new set of training (Step 2) to improve the classifier's performance further.}
\end{description}

This pipeline is typically executed step-by-step, but this is not a rigid constraint. If an expert has a little data, Step 1 is sufficient to speed up its work. Step 1 and Step 2 are helpful in supervised DL-approaches that demand large amounts of highly targeted training data. Finally, since TagLab supports the import of automatic predictions as color-coded images or GIS shapefiles, users can also run an external semantic segmentation model and improve the accuracy of predictions through Step 3. 


\subsection{Comparison with other approaches}

In this section, we review the most popular human-in-the-loop software for marine image data annotation, comparing their workflows with the TagLab one, see Table \ref{tab:comparison}. BIIGLE~\cite{biigle} is a web-based collaborative image and video annotation platform which integrates a Mask R-CNN \cite{MaskRCNN} that can be fine-tuned on custom data through a human-in-the-loop approach detailed in \cite{MAIA2018}. BIIGLE follows a yes/no paradigm for the dataset preparation, proposing a set of automatically segmented regions that users can accept or discard. Additionally, it does not support the editing of predictions. Squidle+ \cite{squidle}, is a cloud-based platform for annotating points in underwater images, which follows an active learning approach. The system has its automatic classifier, and users can feed it with additional inputs to improve its automatic classification performance. However, Squidle+ doesn't allow to modify outliers. Finally, VIAME Toolkit \cite{VIAMEpaper} is an open-source framework for underwater image and video analytics, which includes a Rapid Model Generator feature to create a custom object detector. This approach shares some similarities with TagLab because the automatically detected objects can be edited by manually redrawing incorrect bounding boxes. 

Unlike other software, TagLab's interactive tools allow for the fast creation of a region-labeled training dataset from scratch and the improvement of predictions by using fast per-pixel editing tools. 
Finally, in addition to annotation functionalities, TagLab offers several image analysis options involving low-level information (such areas, centroids, perimeters) and higher-level information such as DEMs or the detection of changes in multi-temporal surveys. 

\begin{table}
  \centering
  \begin{tabular}{cm{3.7cm}m{4.4cm}m{2.9cm}}
  \hline
 Name & Assisted annotation & Fully automatic annotation & Editing  \\
   \hline
  TAGLAB & AI-assisted contours tracing & Guided tuning of custom semantic segmentation models & AI-assisted contours editing \\
  SQUIDLE+ & Active Learning procedure & Point classification performance improves with use & N.a. \\
  BIIGLE & Automated region proposals and validation through Yes/No paradigm & Fine tuning of custom semantic segmentation models & N.a. \\
  VIAME & Detection by examples & Assisted creation of detectors & Manual bounding box editing \\
  \hline 
  \end{tabular}
  \vspace{0.3cm}
  \caption{Comparison of functionalities between human-centric AI annotation software targeted to marine image analysis.}
  \label{tab:comparison}
\end{table}

\subsection{Technical details}

\textbf{Step 1} can be performed using three levels of automation: manual, semi-automatic, and fully automatic.
Regions can be manually outlined through standard drawing and editing tools, such as \emph{Freehand} drawing tool, a pen, and the \emph{Cut} tool, the scissors, to separate segmented areas. Additionally, we designed a  \emph{Edit Border} tool to modify region boundaries with minimal effort, without the need to drag them or to specify any drawing parameters. Users sketch an arbitrary number of intersecting curves, and the inner and outer portions are automatically removed or created. Finally, the \emph{Refinement} tool, which implements a custom version of the graph-cut algorithm~\cite{Boykov2001}, improves the outlining accuracy with the constraint that new segments must remain close to the originals. The \emph{semi-automatic} approaches adopt interactive CNNs for the fast-tracing of boundaries with the minimal cognitive workload. TagLab integrates two agnostic CNN in two tools, the \emph{4-clicks}, based on the Deep Extreme Cut \cite{Maninis2018} and the \emph{positive/negative clicks}, based on the \cite{reviving2021}. The 4-clicks tool segments object by indicating their extremes; clicking the objects' extremes is five-time faster than drawing a bounding box and ensures a higher segmentation accuracy \cite{Papadopoulos2017}. The positive/negative clicks tool enables object outlining by placing a few inner (positive) or outer (negative) points. For less complex shapes, a single point is generally sufficient for achieving complete object outlining. Both CNNs, originally pre-trained on everyday objects, have been fine-tuned on a dataset of 15,000 manually segmented coral colonies, which are, in nature, one of the most shape-complex natural structures. Finally, the orthoimage segmentation can also be obtained in a \emph{fully automatic} way, running an available semantic segmentation model. Since TagLab supports the loading of orthoimages of $32000\times32000$ pixels, predictions are calculated on a sliding window that crops sub-images at a full resolution and re-combines them, weighting probabilities to avoid tiling artifacts. Regardless of how they have been created, labeled regions are displayed as transparent polygons superimposed on the orthoimage. 

\begin{figure}
    \centering
    \includegraphics[height=2.9cm]{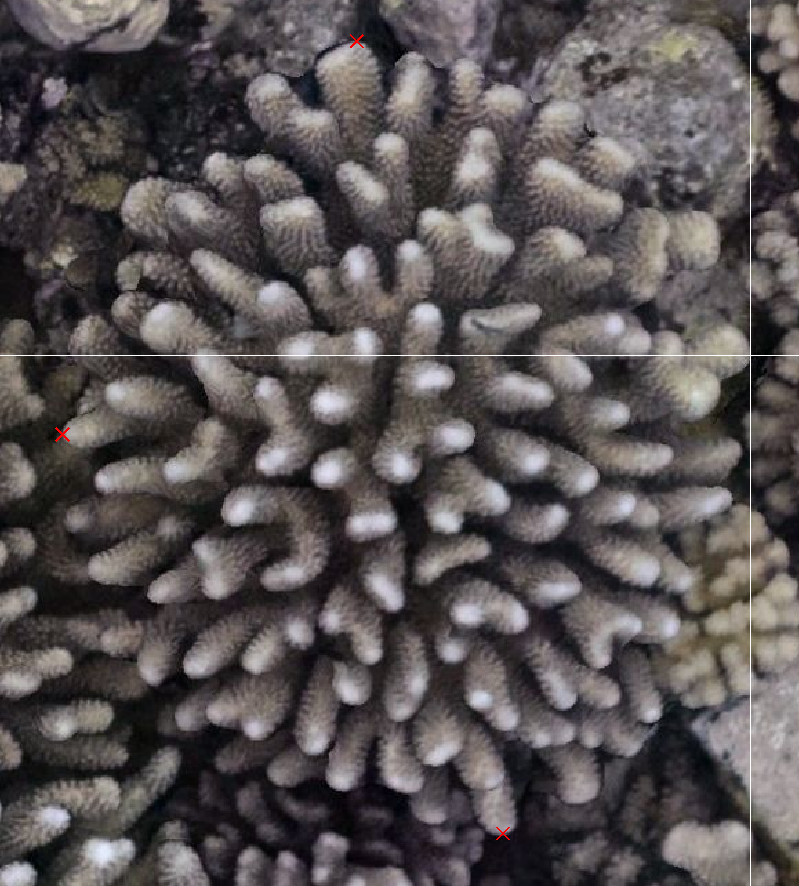}
    \includegraphics[height=2.9cm]{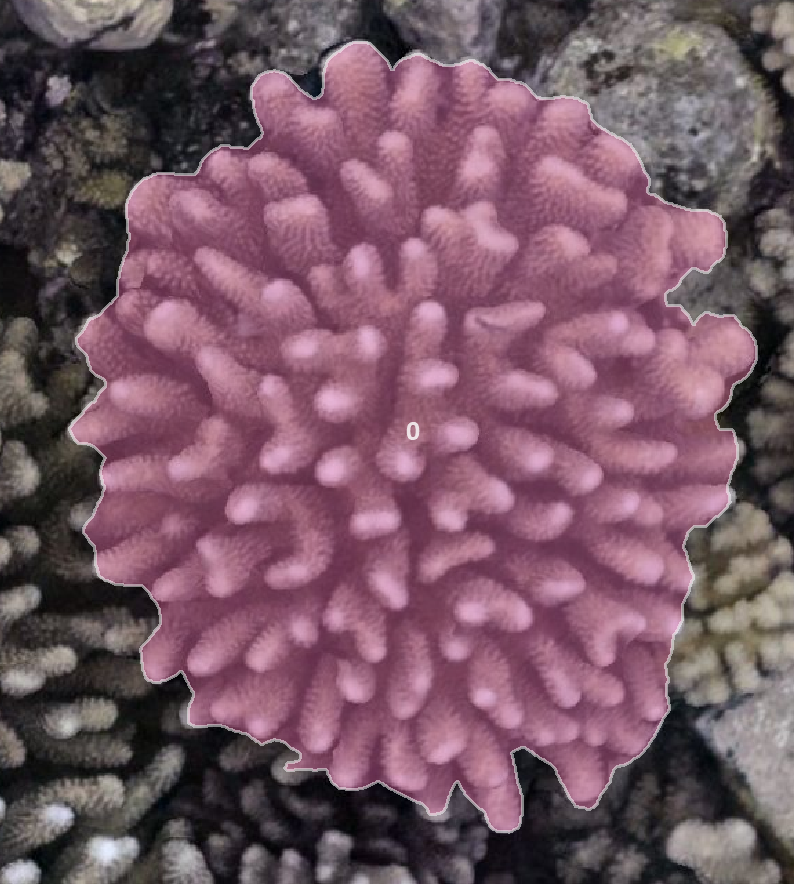}
    \includegraphics[height=2.9cm]{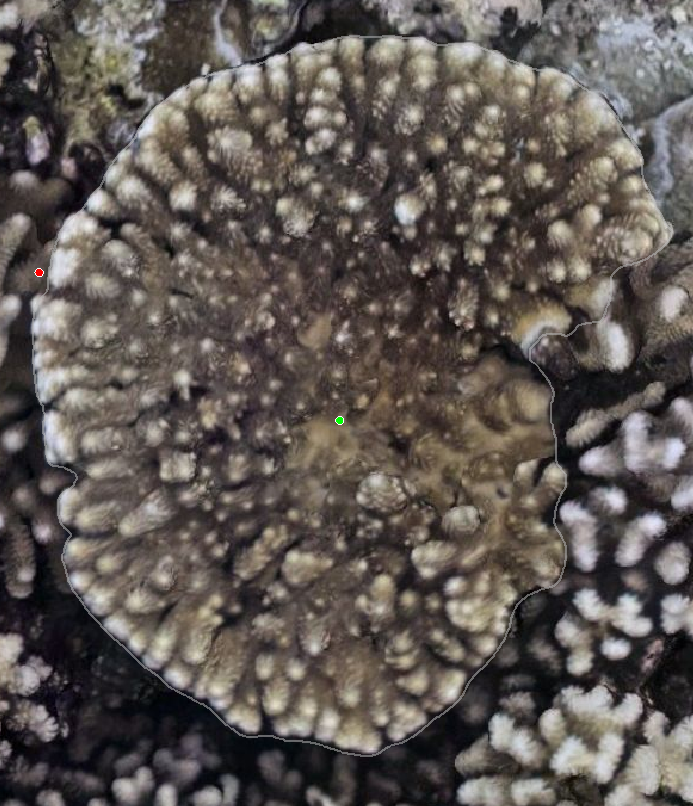}
    \includegraphics[height=2.9cm]{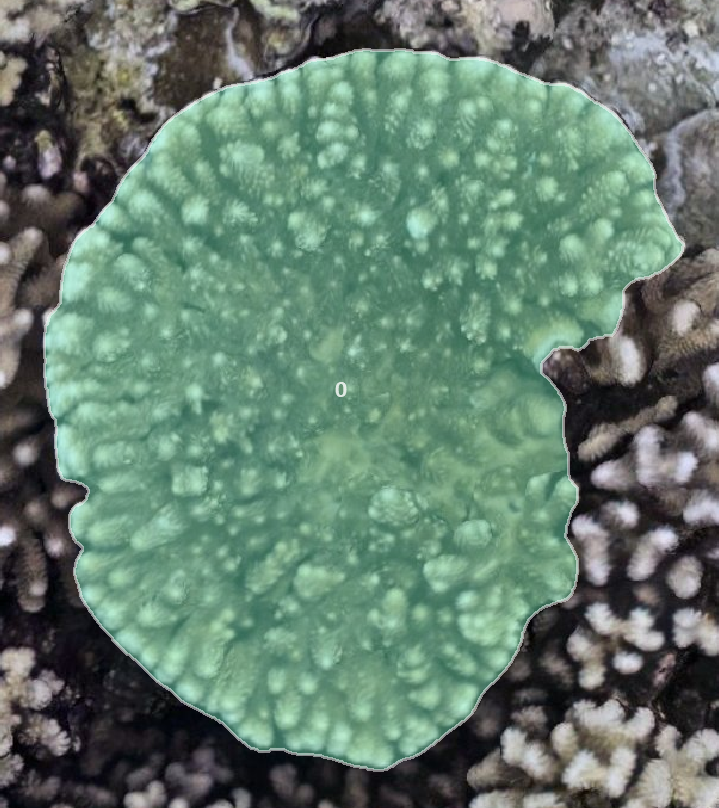}
    \includegraphics[height=2.9cm]{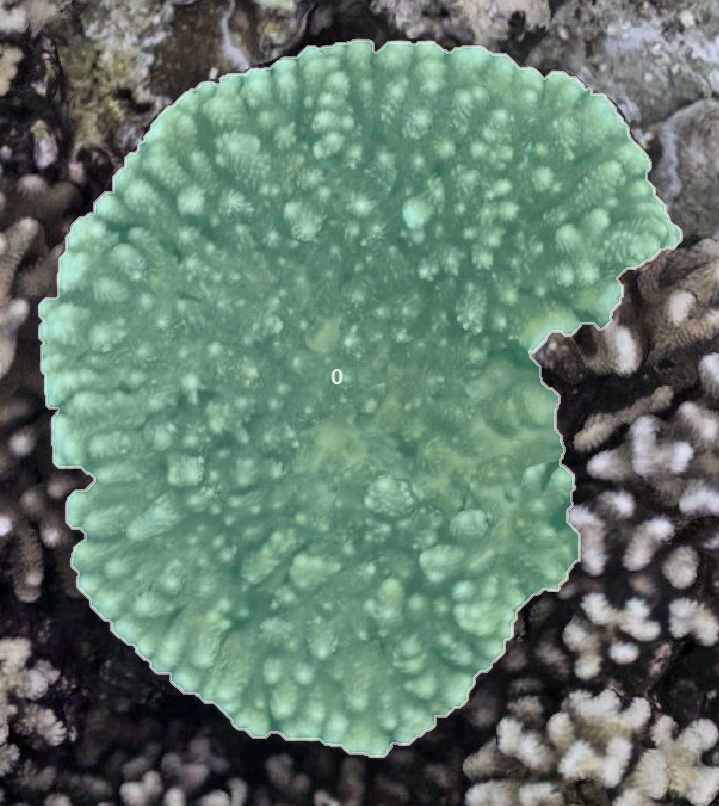}
    \caption{Assisted annotation. From left to right: \emph{4-clicks} tool in action, associated segmentation, \emph{positive/negative click} tool in action, associated segmentation, the use of \emph{Refinement} tool.}
    \label{fig:toolsinaction}
\end{figure}

\textbf{Step 2} On several specific application contexts, there are no existing pre-trained classifiers; thus, users must create custom models. The dataset preparation, training, and validation procedures have been integrated into a guided sub-pipeline which does not require a machine learning background. The Export New Training Dataset functionality facilitates the dataset preparation; a user-friendly interface allows selecting the annotated area of interest on the orthoimage and exports three sets of tiles, for the training, validation, and test, according to different (choosable) criteria. TagLab allows the merging of multiple tiles dataset coherently with the orthoimages scale. Then, users can fine-tune a custom segmentation model starting from a pre-trained CNN, the DeepLab V3+ \cite{deeplab_2018}, through \emph{Train Your Network} function.  Details about the fine-tuning strategies are available in  \cite{Pavoni_2020}. Importantly, TagLab has a set of default hyperparameters that have been chosen after testing the CNN accuracy and generalization capabilities on several datasets. When the fine-tuning ends, the training metrics (accuracy, mIoU, CM) are displayed to evaluate the classifier's performance. Additionally, we added an interface to easily navigate test tiles and relative ground truth masks and predictions (paired). When users find that the automated model produces satisfactory results, they can save it and preview his working on an area of interest on a new orthoimage (before running the fully automatic semantic segmentation on its entire area).

\textbf{Step 3} allow improving the accuracy of automatic predictions through human supervision. Segmented regions can be edited using the annotation tools of Step 1. The editing is straightforward because automatic outputs are visualized as superimposed polygons. When automatic predictions create too smooth regions, the Refinement helps to increase the polygon adherence to the complex shape. The positive/negative clicks tool allows the rapid adjustment of boundaries intuitively, using a few clicks.

\section{Experimental results}

\begin{figure}
    \centering
    \includegraphics[width=0.85\textwidth]{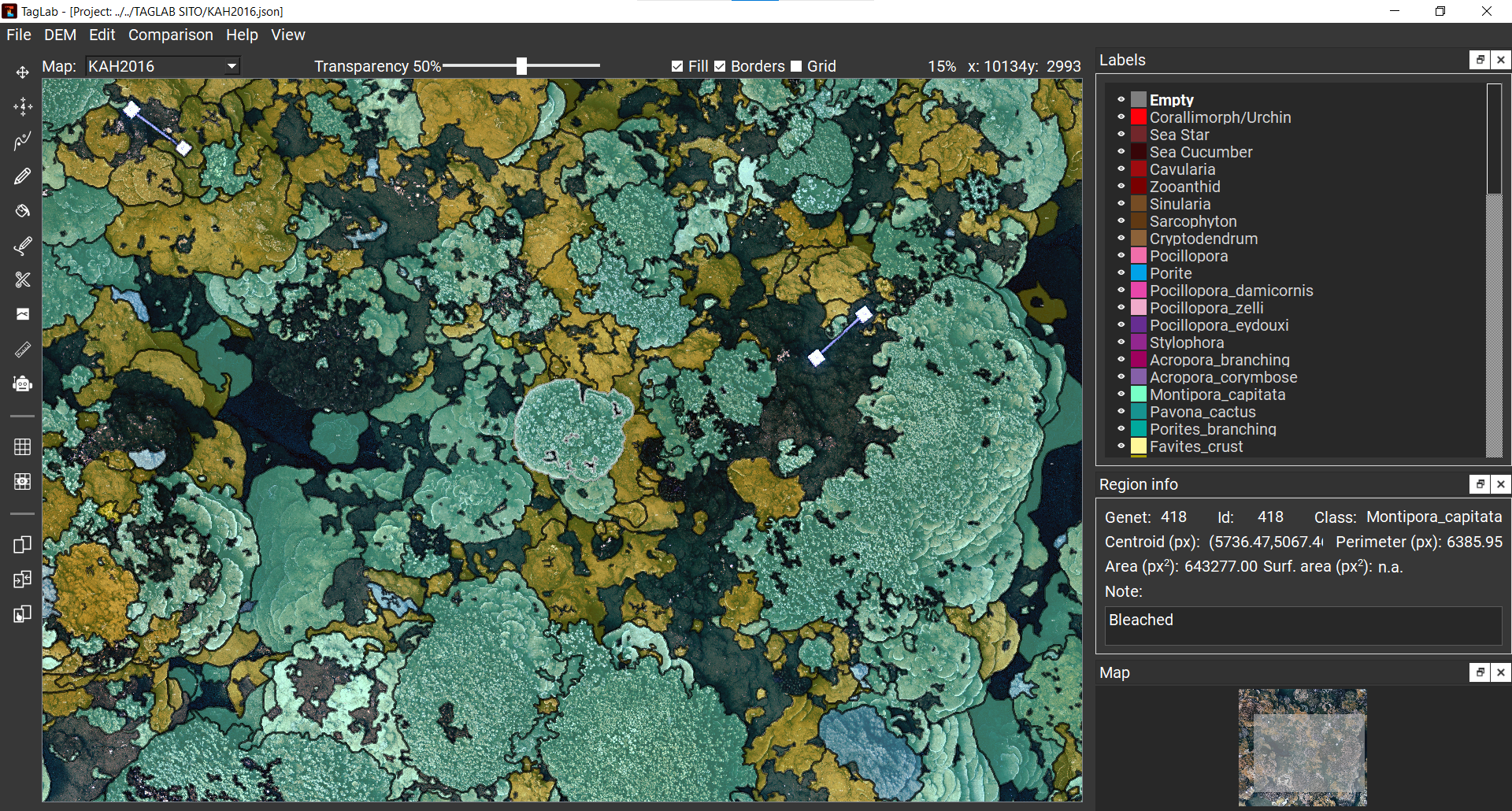}
    \caption{TagLab interface. In the working view, we can appreciate the complexity of coral shapes and the impressive intra-class (same color) morphological variability.}
    \label{fig:interface}
\end{figure}

The \textbf{100 Island Challenge} is a large-scale monitoring project headed by the Sandin Lab from the Scripps Institution of Oceanography (UCSD). Researchers want to assess which factors (human activity, geomorphological conditions, etc.) influence the structure and growth of benthic communities \cite{Ed2017}. To do so, they collect reef orthoimages and perform demographic studies on populations from manually outlined coral taxa (see Fig.\ref{fig:interface}). Their traditional segmentation workflow, performed using Photoshop, takes about 1 hour per sq.m of human work. Together with the Scripps', we conducted a user study to evaluate TagLab's \emph{accuracy} and \emph{efficiency} (in terms of per-pixel tracing speed) of both Step 1 and Step 3. The study was conducted in February 2020 and involved 8 annotators divided in \emph{experts} and \emph{beginners} and 4 orthoimages 
having a different coral coverage and shape complexity. Each orthoimage measured around $3\times3$ meters and had an average pixel size is around 1 mm. At that time, TagLab included as AI-assisted labeling solution only the \emph {4-clicks} tool;  the \emph {positive/negative} clicks tool was added later. The assisted mode increases the annotation speed by 42\% on average and by about 90\% for non-expert annotators, preserving a comparable accuracy on boundaries. The automatic models scored a  per-pixel classification accuracy ranging between 0.8 and 0.98. Due to this variability,  Step 3 showed a lower average speed gain (12\%). However, Step 3 increased the accuracy of automatic outputs by about 7\% on average, with a peak of 14\% for the less accurate predictions. All the annotators improved the training speeds during the user study, suggesting that more confidence with TagLab could improve results. In September 2021, we released a new TagLab version, and we conducted a preliminary study to assess the improvement brought by the\emph{positive/negative} clicks tool. The annotation speed achieved by introducing the new interactive CNN gains 59\% over the old TagLab version and 96\% compared to Photoshop manual workflow  \cite{taglabJFR}.

\textbf{Architectural Heritage: the masonry's interpretation.} Archeologists and engineers use photogrammetric surveys to document and evaluate the status of conservation of masonry walls.  We applied our human-centric AI approach to speed up the annotation work of masonry walls orthoimages in its constituent elements \cite{wall2020}. Homogeneous areas, such as bricks, stones of different types, putlog holes, carpets, etc. are annotated in TagLab, and the working time required is compared with the typical manual workflow used by a group of engineers. The annotation from scratch (Step 1) gets about 40 minutes of work per-orthoimage w.r.t the usual 1 hour and a half. The custom classifier created following all the pipeline steps allows annotating a new orthoimage with excellent accuracy (mIoU greater than 0.96 on the tested orthoimages) in 20 minutes of editing only. In this case, Step 3 proved to be more convenient than the assisted labeling thanks to the excellent performance of the automatic semantic segmentation model.

\section{Conclusions}

Reducing the time required for image post-processing allows researchers from different fields to analyze larger volumes of data. This is fundamental for some disciplines, such as marine ecology, where having a more significant amount of data available means having a greater ability to understand and predict future changes in underwater ecosystems. If, from one side, there is the need to speed up manual annotation, on the other hand, automatic recognition models are not yet as good as humans in performing cognitively complex labeling tasks. 
Here, we described a flexible human-centric AI-based pipeline to speed up the semantic segmentation of orthoimages in these complex scenarios.
We integrated this pipeline into an open-source annotation software, TagLab, that we tested on two real-world case studies. The human-machine collaboration proved to be faster than manual and more accurate than automatic. Furthermore, although automatic recognition via CNN is an established technology, the user-oriented guided learning pipeline of TagLab (Step 2) opens this technology to a broader audience.

\subsubsection*{Acknowledgments}

This work was partially supported by the Italian Minister of University and Research [grant PNRA18\_00263-B2, ``Ross Sea Benthic Monitoring Program: new non-destructive and machine-learning approaches for the analysis of benthos patterns and dynamics''.

\small

\bibliographystyle{eg-alpha-doi}
\bibliography{biblio}

\end{document}